\documentclass[a4paper,english]{scrartcl}
\usepackage[T1]{fontenc}
\usepackage[latin9]{inputenc}
\usepackage{graphicx}

\makeatletter

\date{}
\makeatother
\usepackage{babel}

\begin{document}

\title{Trade-offs in exploiting body morphology for control: from simple bodies and model-based control to complex bodies with model-free distributed control schemes}

\author{
Matej Hoffmann$^1$ and Vincent C. M\"{u}ller$^2$,$^3$ \\ 
\begin{tabular}{cc}
$^1$ iCub Facility, Istituto Italiano di Tecnologia, Genoa, Italy \\
 matej.hoffmann@iit.it\\ 
$^2$ Division of Humanities \& Social Sciences, Anatolia College/ACT, \\
 Pylaia-Thessaloniki, Greece \\
  	vmueller@act.edu \\
$^3$ University of Oxford, UK  
\end{tabular} 
}

\maketitle
\begin{abstract} 
Tailoring the design of robot bodies for control purposes is implicitly performed by engineers, however, a methodology or set of tools is largely absent and optimization of morphology (shape, material properties 
of robot bodies, etc.) is lagging behind the development of controllers. This has become even more prominent with the advent of compliant, deformable or "soft" bodies. These carry substantial potential regarding their exploitation for control---sometimes referred to as "morphological computation" in the sense of offloading computation needed for control to the body. Here, we will argue in favor of a dynamical systems rather than computational perspective on the problem. Then, we will look at the pros and cons of simple vs. complex bodies, critically reviewing the attractive notion of "soft" bodies automatically taking over control tasks. We will address another key dimension of the design space---whether model-based control should be used and to what extent it is feasible to develop faithful models for different morphologies.
\end{abstract}

\section*{Introduction}
It has become increasingly common to explain the intelligent abilities of natural agents through reference to their bodily structure, their morphology, and to make extended use of this morphology for the engineering of intelligent abilities in artificial agents, e.g. robots--thus "offloading" computational processing from a central controller to the morphology. These uses of morphology for explanation and engineering are sometimes referred to as "morphological computation". As we argue in detail elsewhere \cite{Mueller2015}, only some of the characteristic cases that are embraced by the community as instances of morphological computation have a truly computational flavor. Instead, many of them are concerned with exploiting morphological properties to simplify a control task. This has been labeled "morphological control" in \cite{Fuchslin2013}; "mechanical control" could be an alternative label. Developing controllers that exploit a given morphology is only a first step. The space of possible solutions to a task increases dramatically once the mechanical design is included in the design space. At the same time, the dimensionality of the design problem grows exponentially.

In this work, we want to take a close look at these issues. First, we will borrow the "trading spaces" landscape from \cite{Pfeifer2013} that introduces a number of characteristic examples and distributes them along a metaphorical axis from "informational computation" to "morphological computation". Second, we will argue in favor of a dynamical systems rather than computational perspective on the problem. Third, we will critically look at the pros and cons of simple vs. complex (highly dimensional, dynamic, nonlinear, compliant, deformable, "soft") bodies. Fourth, we will address another key dimension of the design space---whether model-based control should be used and to what extent it is feasible to develop faithful models for different morphologies. We will close with an outlook into the future of "soft" robotics. 										

\section*{Design "trading spaces"}

\begin{figure}
	\begin{centering}
		\includegraphics[scale=0.2]{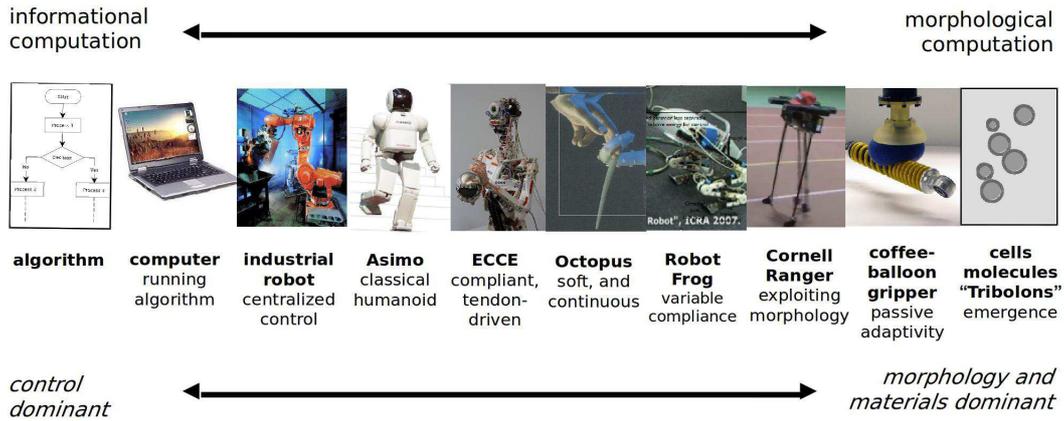}
	\par\end{centering}
	\caption{\textbf{The design trading space.} This figure illustrates the degree to which each system relies on explicit control or self-organization of mechanical dynamics. On the left-hand side of the spectrum, computer algorithms and commercial computers rely on physical self-organization at the minimum level, while towards the right-hand side, more embodied, more soft, and smaller-scale systems require physical interactions as driving forces of behaviors. The design goal then is to find a proper compromise between efficiency and flexibility, taking into account that a certain level of flexibility can also be achieved by changing morphological and material characteristics. (Fig. and caption from \cite{Pfeifer2013}).\label{fig:trading_spaces}}
\end{figure}

Pfeifer et al. \cite{Pfeifer2013} offer one possible perspective on the problem in Fig. \ref{fig:trading_spaces}. In traditional robots---as represented by industrial robots and Asimo in the figure---, control is essentially confined to the software domain where a model of the robot exists and current state of the robot and the environment is continuously being updated in order to generate appropriate control actions sent to the actuators. In biological organisms, on the other hand, this does not seem to be the case: the separation between "controllers" and "controlled" is much less clear and behavior is orchestrated through a distributed network of interactions of informational (neural) and physical processes. Furthermore, there is no centralized neural control, but a multitude of recurrent loops from the lowest level (reflexes and pattern generators in the spinal cord) to different subcortical and cortical areas in the brain. At the same time, the bodies themselves tend to be much more complex in terms of geometrical as well as dynamical properties. This has motivated the design of compliant, tendon-driven robots like ECCE \cite{Wittmeier2013} and soft, deformable robots like Octopus (e.g., \cite{Laschi2012}). However, compared to humans or biological octopus, a comparable level of versatility and robustness in the orchestration of behavior has not yet been achieved in the robotic counterparts. In more restricted settings, the design and subsequent exploitation of morphology is easier, as the jumping and landing robot frog \cite{Niiyama2007}, the passive dynamic based walker, or the coffee balloon gripper demonstrate. The passive dynamic walkers \cite{McGeer1990} are a powerful demonstration that appropriate design of morphology can generate behavior in complete absence of software control. Yet, there is only a single behavior and the environmental niche is very narrow. The coffee-balloon gripper \cite{Brown2010} employs a similar strategy, but achieves surprising versatility on the types of objects that can be grasped. Body designs that follow this guideline were also labeled "cheap designs" \cite{Pfeifer1999}.																										

\section*{From computational to dynamical systems perspective on the control problem}
A general formulation of a control problem in control theory is making a dynamical system follow a desired trajectory. For our purposes, we will consider the cases where the dynamical system is physical---the body of the agent; in control theory, this is the so-called plant. There are numerous control schemes and branches of control theory and the reader is referred to abundant literature on the topic (e.g., \cite{Astrom2008,Emami2002,Kirk2004}). The performance of the controller can be evaluated on various grounds: precision of a trajectory with respect to a reference trajectory, or energy expenditure, for example. In addition, performance, stability and robustness guarantees are required by industry. Control theory typically deals with the design of controllers that optimize these criteria. Some control schemes with appropriate cost functions will automatically result in minimal control actions and thus "optimize the contribution of the morphology". For example, Moore et al. \cite{Moore2012} used Discrete Mechanics and Optimal Control to steer a satellite while exploiting its dynamics to the maximum. Carbajal \cite{Carbajal2012} developed related methods for reaching, plus offered a formalization of the concept of "natural dynamics". Nevertheless, the plant is treated as fixed in these approaches. Yet, the properties of the physical body obviously have a key influence on the final performance of the whole system (plant + controller), which calls for including them into the design space. Therefore, including the body into the design space carries great potential.

We feel that to address the problem of morphology simplifying control, a computational perspective does not provide the most appropriate framework. Instead, the dynamical systems description seems more appropriate for the following reasons: (i) It fits the informational and physical processes equally well; (ii) It copes with continuous (in time) streams of continuous input and output signals; (iii) It is already used by control theory. The concept of self-stabilization, for example, which is often cited in the "morphological computation community" is naturally explained from a dynamical systems standpoint.

In the case of the passive dynamic walker or the jamming-based grippers cited above, the body is ingeniously contributing to its, perhaps primary, function: enabling physical behavior in the real world. While this is often interpreted in the "offloading sense"---the body design takes over computation from the brain (e.g., \cite{PfeiferBongard2007})---we argue that the contribution of the body itself (and the interaction with the environment) is not computational in any substantial sense. In fact, calling it computation may allude to a much stronger thesis (at least in the classical notion of computation), namely that what is done by the interaction here could also be done by a conventional computer. And that is quite the opposite of what most proponents of the morphological stance want to say. Therefore, our claim is that exploiting morphology for control should be better handled in a non-computational context and that the dynamical systems framework is more appropriate than a classical computational one. 

\section*{Simple or complex bodies?}
The spirit of the morphological computation literature that follows the "offloading" or "trade-off" perspective, is that complex (highly dimensional, dynamic, nonlinear, compliant, deformable, "soft" ) bodies are advantageous for control because they can take over the "computation" that a controller would otherwise have to perform (e.g., \cite{Hauser2011,Hauser2012,PfeiferBongard2007} or  \cite{Caluwaerts2013} explicitly in Fig. 1). Complex nonlinear bodies give rise to more complex dynamical landscapes where the location of attractors can facilitate the performance on a given task. 

This view is in stark contrast to the views prevalent in control theory. There, linear time-invariant systems are the ideal plants to control. Solutions for nonlinear systems are much more difficult to obtain and they often involve a linearization of the system of some sort. In fact, human-like bodies are a nightmare for control engineers (\cite{Potkonjak2011} is an interesting case study) and highly complex models and controllers would be required. 

What would be an ideal body then? And, does a complex body imply simple or complex control? Recent attempts at quantifying the amount of morphological computation shed more light on this issue. Zahedi and Ay \cite{Zahedi2013} propose two concepts for 
measuring the amount of morphological computation by calculating the conditional dependence of future world states (or body state) on previous world states and action taken by the agent. According to Concept 1, the amount of morphological computation is 
inversely proportional to the contribution of the agent's actions to the overall behavior. Concept 2 equates the amount of morphological computation with the contribution of the world to the overall behavior. From a robotic perspective, optimizing for morphological computation, Concept 1 is equivalent to systems that cannot be controlled by actions of the robot; Concept 2 will give rise to systems with strong "body dynamics" or "natural dynamics" (see e.g., \cite{Iida2005} or \cite{Carbajal2012} for a formal definition). A body with strong internal dynamics (Concept 2), resisting control actions (Concept 1), will not be a product of typical engineering designs. Yet, if it is possible to design the body to assist in task performance, it is certainly of advantage. The passive dynamic walker is an extreme case; however, underactuated designs featuring active and passive degrees of freedom are also abundant.

R\"{u}ckert and Neumann \cite{Ruckert2013} study learning of optimal control policies for a simulated 4-link pendulum which needs to maintain balance in the presence of disturbances. The morphology (link lengths and joint friction and stiffness) is manipulated and controllers are learned for every new morphology. They show that: (1) for a single controller, the complexity of the control (as measured by the "variability" of the controller) varies with the properties of the morphology: certain morphologies can be controlled with simple controllers; (2) optimal morphology depends on the controller used; (3) more complex (time-varying) controllers achieve much higher performance than simple control across morphologies. 

In summary, the performance on a task will always depend on a complex interplay of the controller, body, and environment: taking out the controller is just as big a mistake as taking out the body was. The tasks that can be completely solved by appropriate tuning of the body, like passive dynamic walking, are the exception rather than the rule. A controller will thus be needed too. A complex body may have the potential to partially solve certain tasks on its own; yet, it may present itself as difficult to control, model (if the controller is relying on models), design, and manufacture. An optimal balance thus needs to be found. For that, however, new design methodologies that would encompass complex cost functions (performance on a task, versatility, robustness, costs associated with hardware whose parameters can be manipulated etc.) are needed. Hermans et al. \cite{Hermans2014} very recently proposed such a method that uses machine learning to optimize physical systems; an approximate parametric model of the system's dynamics and sufficient examples of the desired dynamical behavior need to be available though---which leads us to the next section.

\section*{With or without a model?}
Including the parameters of the body into the design considerations may give rise to better performance of the whole system; these may be solutions involving a simpler controller, but also solutions that were previously unattainable when the body was fixed. Following the dynamical system's perspective, \cite{Fuchslin2013} provide an illustration of the possible goals of the design process: (1) To design the physical dynamical system such that desired regions of the state space have attracting properties. Then it is sufficient to use a simple control signal that will bring the system to the basins of attraction of individual stable points that correspond to target behaviors. (2) More complicated behavior can be achieved if the attractor landscape can be manipulated by the control signal. 

If a mathematical formulation of the controller and the plant is available, this design methodology can be directly applied. The first part is demonstrated by McGeer \cite{McGeer1990} on the passive dynamic walker: The influence of scale, foot radius, leg inertia, height of center of mass, hip mass and damping, mass offset, and leg mismatch is evaluated. In addition, the stability of the walker is calculated. Recently, Jerrold Marsden and his coworkers presented a method that allows for co-optimization of the controller and plant by combining an inner loop (with discrete mechanics and optimal control) and an outer loop (multiscale trend optimization). They applied it to a model of a walker and obtained the best position of the knee joints (\cite{Pekarek2010} -- Ch. 5).

However, typical real-world agents are more complex than simple walkers. Holmes et al. \cite{HolmesFull2006} provide an excellent dynamical systems analysis of the locomotion of rapidly running insects and derive implications for the design of the RHex robot. Yet, they conclude that "a gulf remains between the performance we can elicit empirically and what mathematical analyses or numerical simulations can explain. Modeling is still too crude to offer detailed design insights for dynamically stable autonomous machines in physically interesting settings." Hermans et al. \cite{Hermans2014} similarly note that applying their method to robotics, which is known to suffer from lack of accurate models, is a challenge. The modeling and optimization of more complicated morphologies---like compliant structures---is nevertheless an active research topic (e.g., Wang \cite{Wang2009} and other work by the author). The second point of F\"{u}chslin et al. \cite{Fuchslin2013}---achieving "morphological programmability" by constructing a dynamical system with a parametrized attractor landscape---remains even more challenging though. 

One of the merits of exploiting the contributions of body morphology should be that the physical processes do not need to be modeled, but can be used directly. However, without a model of the body at hand, several body designs need to be produced and---together with the controller---tested in the respective task setting. The design space of the joint controller-body system blows up and we may be facing a curse of dimensionality. This is presumably the strategy adopted by the evolution of biological organisms that could cope with the enormous dimensionality of the space. In robotics, this has been taken up by evolutionary robotics \cite{Nolfi2000}.  The simulated agents of Karl Sims \cite{Sims1994} demonstrate that co-evolving brains and bodies together can give rise to unexpected solution to problems. More recently, with the advent of rapid prototyping technologies, physics-based simulation could be complemented by testing in real hardware \cite{Lipson2000}. Yet, a "reality gap" always remains between simulated and real physics and searching for optimal designs in hardware directly is very costly (e.g., \cite{Hornby2000}). The design decisions---which parameters to optimize---are based on heuristics and a clear methodology is still missing. Furthermore, with the absence of an analytical model of the controller and plant, no guarantees on the system's performance can be given.  

\section*{Outlook}
In terms of applications, the most relevant area where exploitation of morphology is and will be the key is probably robotics, and in particular soft robotics (see \cite{AlbuSchaffer2008,Pfeifer2012,Pfeifer2013,Trivedi2008} and the first issue of the Soft Robotics Journal \cite{Trimmer2013}). "Soft" robots break the traditional separation of control and mechanics and exploit the morphology of the body and properties of materials to assist control as well as perceptual tasks. Pfeifer et al. \cite{Pfeifer2012} even discuss a new industrial revolution. Appropriate, "cheap", designs lead to simpler control structures, and eventually can lead to technology that is cheap in a monetary sense and thus more likely to impact on practical applications. Yet, a lot of research in design, simulation and fabrication is needed (see \cite{Lipson2013} for a review).   

The area of soft robotics and morphological computation seems to be rife with different trading spaces \cite{Pfeifer2013}. As we move from the traditional engineering framework with a central controller that commands a "dumb" body toward delegating more functionality to the body, some convenient properties will be lost. In particular, the solutions may not be portable to other platforms anymore, as they will become dependent on the particular morphology and environment (the passive dynamic walker is the extreme case). The versatility of the solutions is likely to drop as well. To some extent, the morphology itself can be used to alleviate these issues---if it becomes adaptive. Online changes of morphology (like changes of stiffness or shape) thus constitute another tough technological challenge (see also project LOCOMORPH \cite{Locomorph}). Completely new, distributed control algorithms that rely on self-organizing properties of complex bodies and local distributed control units will need to be developed (the tensegrity structure controlled by a  spiking neural network \cite{Rieffel2010} is a step in this direction). 

\section*{Acknowledgements}
M.H. was supported by the Swiss National Science Foundation Fellowship PBZHP2-147259. M.H. also thanks Juan Pablo Carbajal for fruitful discussions and pointers to literature. Both authors thank the EUCogIII project (FP7-ICT 269981) for making us talk to each other.

\newpage
\bibliography{HoffmannMueller_MorphControl_e_book}
\bibliographystyle{plain}

\end{document}